\documentclass[sn-chicago]{sn-jnl}

\usepackage{graphicx}%
\usepackage{multirow}%
\usepackage{amsmath,amssymb,amsfonts}%
\usepackage{amsthm}%
\usepackage{mathrsfs}%
\usepackage{stmaryrd,nicefrac}%
\usepackage[title]{appendix}%
\usepackage{xcolor}%
\usepackage{textcomp}%
\usepackage{manyfoot}%
\usepackage{booktabs}%
\usepackage{algorithm}%
\usepackage{algorithmicx}%
\usepackage{algpseudocode}%
\usepackage{listings}%
\usepackage{comment}
\usepackage[bottom]{footmisc}

\newcommand*{\defeq}{\mathrel{\vcenter{\baselineskip0.5ex \lineskiplimit0pt
      \hbox{\footnotesize.}\hbox{\footnotesize.}}}%
  =}

\newcommand{\fromto}{\longrightarrow}

\renewcommand{\vec}[1]{\boldsymbol{#1}}

\DeclareMathOperator{\prob}{\mathit{p}}

\newcommand{\cD}{\mathcal{D}}

\newcommand{\given}{\mid}

\newcommand{\scoresys}[3]{\ensuremath{\langle #1,#2,#3 \rangle}}

\renewcommand{\cite}[1]{\citep{#1}}

\theoremstyle{thmstyleone}%
\theoremstyle{thmstyletwo}%

\theoremstyle{thmstylethree}%
\newtheorem{definition}{Definition}%

\begin{document}

\title[Article Title]{Probabilistic Scoring Lists for Interpretable Machine Learning}


\author*[1,2]{\fnm{Jonas} \sur{Hanselle}}\email{jonas.hanselle@ifi.lmu.de}
\equalcont{These authors contributed equally to this work.}
\author*[1,2]{\fnm{Stefan} \sur{Heid}}\email{stefan.heid@ifi.lmu.de}
\equalcont{These authors contributed equally to this work.}
\author*[3]{\fnm{Johannes} \sur{Fürnkranz}}\email{juffi@faw.jku.at}
\author*[1,2]{\fnm{Eyke} \sur{Hüllermeier}}\email{eyke@lmu.de}

\affil*[1]{\orgdiv{Institute of Informatics}, \orgname{LMU Munich}, \orgaddress{\street{Akademiestr.\ 7}, \city{Munich}, \postcode{80799},  \country{Germany}}}

\affil[2]{\orgname{Munich Center for Machine Learning}, \city{Munich}, \country{Germany}}

\affil[3]{\orgdiv{Institute for Application-oriented Knowledge Processing}, \orgname{Johannes Kepler Universität Linz}, \orgaddress{\street{Altenberger Straße 69}, \city{Linz}, \postcode{4040}, \country{Austria}}}


\abstract{A scoring system is a simple decision model that checks a set of features, adds a certain number of points to a total score for each feature that is satisfied, and finally makes a decision by comparing the total score to a threshold. Scoring systems have a long history of active use in safety-critical domains such as healthcare and justice, where they provide guidance for making objective and accurate decisions. Given their genuine interpretability, the idea of learning scoring systems from data is obviously appealing from the perspective of explainable AI. In this paper, we propose a practically motivated extension of scoring systems called probabilistic scoring lists (PSL), as well as a method for learning PSLs from data. Instead of making a deterministic decision, a PSL represents uncertainty in the form of probability distributions, or, more generally, probability intervals. Moreover, in the spirit of decision lists, a PSL evaluates features one by one and stops as soon as a decision can be made with enough confidence. To evaluate our approach, we conduct a case study in the medical domain. }



\maketitle

\section{Introduction} 
\label{sec:intro}

Predictive models generated by modern machine learning algorithms, such as deep neural networks, tend to be complex and difficult to comprehend, and may not be appropriate in applications where a certain degree of transparency of a model and explainability of decisions are desirable. Besides, depending on the situation and application context, time and computational resources for applying decision models might be limited. For example, a human's resources to collect, validate, and enter data might be scarce, or decisions must be taken quickly, in the extreme case even by the human herself without any technical device.  


So-called \emph{scoring systems} provide a simple, genuinely interpretable model class as an alternative. In a nutshell, a scoring system is a decision model that checks a set of features, adds (or subtracts) a certain number of points to a total score for each feature that is satisfied, and finally makes a decision by comparing the total score to a threshold. Scoring systems have a long history of active use in safety-critical domains such as healthcare \citep{six_cp08} and justice \citep{wang_ip22}, where they provide guidance for making objective and accurate decisions. Given their genuine interpretability, scoring systems are appealing from the perspective of explainable AI, which is why the idea of learning such systems from data has recently attracted attention in machine learning. 

Building on our previous work \citep{psl}, this paper contributes to existing methodology for scoring systems as follows:
\begin{itemize}
\item We propose a practically motivated extension of scoring systems called \emph{probabilistic scoring lists} (PSL), as well as a method for learning PSLs from data. 
\item To increase uncertainty-awareness, a PSL produces predictions in the form of probability distributions (instead of making deterministic decisions). 
\item Moreover, to increase cost-efficiency, a PSL is conceptualized as a \emph{decision list}: It evaluates features one by one and stops as soon as a decision can be made with enough confidence. 
\end{itemize}

\noindent
Moreover, we extend \citep{psl} in various ways: 
\begin{itemize}
\item We make the PSL method amenable to continuous variables by developing \emph{discretization techniques} for turning numerical attributes into binary features, either in a preprocessing step or progressively in the course of the PSL procedure. 
\item In order to calibrate probability estimates, we consider beta calibration in addition to isotonic regression.
\item We propose a method for quantifying \emph{epistemic uncertainty}, i.e., uncertainty about the probability estimates in a PSL.
\item Going beyond simple decisions, we propose a variant of PSL that is appropriate for the task of \emph{ranking}, i.e., sorting a set of instances from most likely positive to most likely negative, leveraging training data in the form of relative comparisons.
\item We also expand the empirical evaluation by adding additional datasets and experimental studies. 
\end{itemize}

Following a brief overview of related work in the next section, we introduce PSLs in Section~\ref{sec:scoring} and address the problem of learning such models from data in Section~\ref{sec:learning_psl}. To evaluate our approach, we conduct as series of experimental studies in Section~\ref{sec:exp}, prior to concluding the paper with an outlook on extensions and future work in Section~\ref{sec:future_work}.

\section{Related Work}

In a series of papers, Ustun and Rudin developed the so-called Supersparse Linear Integer Model (SLIM) for inducing scoring systems from data, as well as an extension called RiskSLIM \citep{ustun_supersparse_2016,ustu_or17,ustu_lo19}. Their methods are based on formalizing the learning task as an integer linear programming problem, with the objective to find a meaningful compromise between sparsity (number of variables included) and predictive accuracy. The problem can then essentially be tackled by means of standard ILP solvers.
 
In several applied fields, one also finds methods of a more heuristic nature. Typically, standard machine learning methods such as support vector machines or logistic regression are used to train a (sparse) linear model, and the real-valued coefficients of that model are then turned into integers, e.g., through rounding or by taking the sign.  Obviously, approaches of that kind are rather ad-hoc, and indeed, can be shown to yield suboptimal performance in practice \citep{subr_da21}. From a theoretical perspective, certain guarantees for the rounded solutions can nevertheless be given \citep{chev_rm13}.

A related research direction is the learning of simple decision heuristics that are considered plausible from the perspective of cognitive psychology. Again, however, this is a relatively unexplored field, in which only a few publications can be found so far\,---\,\citet{simsek_learning_2017} collect and empirically compare some of these heuristics. 



Decision lists have been primarily used in inductive rule learning \citep{jf:Book-Rules}, where each term consists of a conjunction of conditions, which are sufficient to make a prediction in case the conditions are satisfied, or else continue with the next rule. They have been shown to generalize both, $k$-term CNF and DNF expressions, as well as decision trees with a fixed depth $k$ \citep{DecisionLists}.
Practically, they represent a simple way for tie-breaking in situations where multiple rules cover the same example: in that case, the first rule in the list is given priority.
They can be easily learned, as their structure mirrors the commonly covering or separate-and-conquer strategy \citep{jf:AI-Review}, which learns one rule at a time, typically by appending rules to the list, assuming that most important rules are tried first, but prepending has also been tried \citep{Prepend}.
While rules are typically used for classification, they may also be viewed as simple probability estimators, using the class distribution among the covered examples as the basis for various estimation techniques \citep{jf:DS-09}.
However, these are known to be overly optimistic, because the way the conditions are selected results in a bias towards the positive examples during learning \citep{ExtremeValueCorrection}.
Also, in decision lists in rule learning, the probability estimates are derived from the last rule in isolation, practically ignoring all previous rules, whereas, as will be seen later, the probability distributions in PSLs are successively refined.

\section{Probabilistic Scoring Lists}
\label{sec:scoring}

Consider a scenario where decisions need to be made in different contexts, which are characterized in terms of a set of variables or features $\mathcal{F} = \{ f_1, \ldots , f_K \}$. A concrete situation is specified by a vector $\vec{x} = (x_1, \ldots , x_K)$, where $x_i$ is the value observed for the feature $f_i$, and the set of all conceivable vectors of that kind forms the instance space $\mathcal{X}$. 
Features can be of various kinds, i.e., binary, (ordered) categorical, or numeric. Decisions are taken from a decision space $\mathcal{Y}$, which is normally finite, typically comprising a small to moderate number of alternatives to choose from.

A decision model is a mapping $h: \mathcal{X} \fromto \mathcal{Y}$, i.e., $y = h(\vec{x})$ is the decision suggested by $h$ in the context $\vec{x}$.
Note that such models can be represented in different ways.
For the reasons already explained, we shall focus on scoring systems in this paper.
In a nutshell, scoring systems consist of a set of simple criteria (presence or absence of certain characteristics or features) that are checked, and if satisfied, contribute a certain number of points to a total score.
The final decision is then based on comparing this score to one or more thresholds.
Formally, scoring systems can be seen as a specific type of generalized additive models \citep{hastie2017} defined over a set of features.

\medskip

\begin{definition}[Scoring system]
A \emph{scoring system} over a set of (binary) candidate features $\mathcal{F}$ and score set $\mathcal{S} \subset \mathbb{Z}$ is a triple $h = \scoresys{F}{S}{t}$, where  $F = \{ f_1, \ldots , f_K \} \subset \mathcal{F}$ is a subset of the  candidate features, $S = (s_1, \ldots , s_K ) \in \mathcal{S}^K$ are scores assigned to the corresponding features, and $t \in \mathbb{Z}$ is a decision threshold.  For a given decision context  $\vec{x} = (x_1, \ldots , x_K) \in \{ 0,1 \}^K$, i.e., the projection of an instance to the feature set $F$, the decision prescribed by $h$ is  given by 
\begin{equation}
h( \vec{x} ) =  \big\llbracket \, T(\vec{x}) \geq t \, \big\rrbracket  = 
\left\llbracket \, \sum_{i=1}^K s_i \, x_i \geq t \, \right\rrbracket \, ,
\label{eq:scoring-system}
\end{equation}
 where $\llbracket \cdot \rrbracket$ is the indicator function.\footnote{$\llbracket P \rrbracket = 1$ if predicate $P$ is true (positive decision) and $\llbracket P \rrbracket = 0$ if $P$ is false (negative decision).}
\end{definition}

\medskip


Note that, according to this definition, scoring systems are binary classifiers ($\mathcal{Y} = \{ 0, 1\}$). In the following, we generalize such scoring systems in two ways: from deterministic to probabilistic, and from a single decision model to a decision list. 

As for the first extension, the idea is to return a probability distribution over $\mathcal{Y}$ instead of a binary decision (\ref{eq:scoring-system}), i.e., to assign a probability $\prob(y)$ to each decision $y \in \mathcal{Y}$. The latter can be interpreted as the probability that $y$ is the best or correct decision, which (implicitly) presupposes the existence of a kind of ground truth. Without loss of generality, we can assume that the ground truth distinguishes between a class of positive cases and a class of negative cases, and that the decision is a prediction of the correct class. Therefore, we shall use the terms ``decision'' and ``class'' interchangeably. 

We contextualize the distribution $\prob$, not directly with $\vec{x}$, but rather with the total score $T(\vec{x})$ assigned to $\vec{x}$. In other words, we consider conditional probabilities $\prob( \cdot \, | \, T(\vec{x}))$ on $\mathcal{Y}$. This appears meaningful and is in line with the assumption that the total score is indicative of the class\,---\,in fact, standard scoring systems can be seen as a special case, returning probability 1 for the positive class when exceeding the threshold and probability 0 otherwise. 

\medskip

\begin{definition}[Probabilistic scoring system, PSS]
A \emph{probabilistic scoring system} (PSS) over candidate features $\mathcal{F}$ and score set $\mathcal{S} \subset \mathbb{Z}$ is a triple $h = \scoresys{F}{S}{q}$, where  $F = \{ f_1, \ldots , f_K \} \subset \mathcal{F}$, $S = (s_1, \ldots , s_K ) \in \mathcal{S}^K$, and $q$ is a mapping $\Sigma \fromto [0,1]$,
where 
$$
\Sigma \defeq \left\{ T = \sum_{i=1}^K s_i \,  x_i \; \bigg| \; s_1, \dots , s_K \in \mathcal{S}, \, x_1, \dots , x_K \in \{ 0, 1 \} \right\}
$$ 
is the set of possible values for the total score that can be obtained by any instance $\vec{x} \in  \mathcal{X}$, 
and $q(T) = \prob( y=1 \given T)$ is the (estimated) probability for the positive class $(y = 1)$ given that the total score is $T$ (and hence $1- q(T) = \prob( y=0 \given T)$ the probability for the negative class).
\end{definition}

Note 
that an increase in the total score should only increase but not decrease the probability of the positive decision, so that probabilistic scoring systems should satisfy the following monotonicity constraint: 
\begin{equation}\label{eq:mono}
\forall \, T , T' \in \Sigma : \, (T < T') \Rightarrow q(T) \leq q(T') \, .
\end{equation}
This property is again in line with standard scoring systems and appears to be important from an interpretability perspective: A violation of (\ref{eq:mono}) would be considered as an inconsistency and compromise the acceptance of the decision model. Therefore, in the remainder of the paper, we consider only monotonic probabilistic scoring systems.

Our second extension combines probabilistic scoring systems with the notion of decision lists.
The underlying idea is as follows:
Instead of determining all $K$ feature values $x_i$ right away, these values are determined successively, one after the other, in a predefined order.
Each time a new feature is added, the total score $T$ is updated, and the probability $q(T)$ of the positive class is determined. Depending on the latter, the process is then continued or stopped:
If the probability is sufficiently high or sufficiently low, the process is stopped, because a decision can be made with enough confidence;
otherwise, the process is continued by adding the next feature. 

\begin{table}[b]
\centering
\caption{Example of a PSL with feature set $\mathcal{F} = \{ f_1, f_2, f_3, f_4 \}$ and score set $\mathcal{S} = \{ 0, \pm 1 , \pm 2 \}$.}
\label{tab:epsl}
\begin{tabular}{ccc@{\ \ \ }c@{\ \ \ }c@{\ \ \ }c@{\ \ \ }c@{\ \ \ }c@{\ \ \ }c}
\toprule
Feature&	Score &	$T \!=\!  -2$ &	$T \!=\!  -1$ & 	$T\!=\!  0$  & $T \!=\!  +1$ &	$T \!=\!  +2$ 	&$T \!=\!  +3$& 	$T \!=\!  +4$ \\
\midrule
$-$  	& $-$ 	 &$-$&  	 $-$ &  	   $0.3$	& $-$  	& $-$  	& $-$ & 	 $-$  \\
 $f_3$ & $\llap{+}1$ 	& $-$ &  	 $-$ &   	   $0.2$ 	& $0.4$ 	& $-$   	& $-$   	& $-$   \\
 $f_1$ & $\llap{$-$}2$ 	& $0.1$ & 	 $0.2$ &  	   $0.5$  	& $0.6$  	& $-$    &	 $-$    	& $-$    \\
 $f_2$ & $\llap{+}1$ 	& $0.1$ &	 $0.2$ & 	   $0.6$ 	& $0.7$ 	& $0.9$ 	& $-$   	 &$-$   \\
 $f_4$	& $\llap{+}2$	 &$0.1$	 &$0.1$ &	       $0.2$	& $0.6$	 &$0.7$	 &$0.9$	& $0.9$\\
\bottomrule
\end{tabular}

\end{table}

Table \ref{tab:epsl} depicts a PSL with four features $F = \{ f_1, f_2, f_3, f_4 \}$.
As can be seen from the assigned scores, all features except $f_1$ are indicative of the positive class, i.e., the presence of $f_2$, $f_3$ or $f_4$ increases the probability of the positive class, whereas the presence of $f_1$ decreases the probability.  

The decision process starts with an empty feature set and a prior probability of $0.3$ for the positive class. After seeing the first feature $f_3$ with a weight $s_3 = +1$, the possible scores are $T = 0$ if the feature does not hold (the value of the feature is $x_3 = 0$), or $T = +1$, if $x_3 = 1$. In the former case, the probability for the positive decision decreases to $0.2$, in the latter case it increases to $0.4$. The next feature is $f_1$ with a weight of $s_1 = -2$, resulting in a total of four possible scores, ranging from $T = -2$ (if $x_3 = 0$ and $x_1 = 1$) to $T = +1$ (if $x_3 = 1$ and $x_1 = 0$). Note that the absence of $f_3$, in this case, may increase the probability of a positive score to $0.6$. Adding the remaining features continues this process, until we get a diverse set of seven probability estimates (five of which are different) corresponding to the seven different score values we can obtain for the $2^4 = 16$ possible instances. For example, the instance $\vec{x} = (1,1,1,1)$ would be assigned a probability of $q(2) = 0.7$, based on its total score of $T(\vec{x}) = +2$.

\color{black}
Note the monotonicity~\eqref{eq:mono} in the scores in each row (higher score values result in higher probabilities for the positive decision).
\color{black}
Also note that if the final maximal probability of $0.9$ is considered to be sufficiently high for making a positive decision, the process could already have been stopped after seeing the first three features for any instance $\vec{x} = (0,1,1,*)$, irrespective of its value $x_4$ for the fourth feature $f_4$.

Formally, we can define a probabilistic scoring list as follows:

\medskip

\begin{definition}[Probabilistic scoring list, PSL]
A \emph{probabilistic scoring list} over candidate features $\mathcal{F}$ and score set $\mathcal{S} \subset \mathbb{Z}$ is a triple $h = \scoresys{F}{S}{p}$, where  $F = ( f_1, \ldots , f_K )$ is a list of (distinctive) features from $\mathcal{F}$, $S = (s_1, \ldots , s_K ) \in \mathcal{S}^K$, and $q$ is a mapping 
\begin{equation}\label{eq:ppsl}
q: \, \bigcup_{k = 0}^K ( k, \Sigma_k)  \, \fromto [0,1] 
\end{equation}
such that 
\begin{equation}\label{eq:mono2}
\forall \, k \in \{0, 1, \ldots , K \}, T , T' \in \Sigma_k : \, (T < T') \Rightarrow q(k,T) \leq q(k,T') \, .
\end{equation}
Here, $\Sigma_k$ is the set of possible values for the total score at stage $k$, i.e., 
\begin{equation}\label{eq:sigmak}
\Sigma_k = \left\{ T = \sum_{i=1}^k s_i \,  x_i \; \bigg| \; s_1, \ldots , s_k \in \mathcal{S}, \, x_1, \ldots , x_k \in \{ 0, 1 \} \right\} \, .
\end{equation}
A value $q(k,T)$ is interpreted as the probability of the positive decision if the total score at stage $k$ is given by $T$. 
\end{definition}

\medskip

Note that $k=0$ is included in (\ref{eq:ppsl}). This case corresponds to the empty list, where no feature has been determined at all. The corresponding value $q(0,0)$ can be considered as a default probability of the positive class.

\section{Learning Probabilistic Scoring Lists}
\label{sec:learning_psl}

While standard scoring systems have often been handcrafted by domain experts in the past, more recent methods for the data-driven construction of scoring systems aim to achieve a good trade-off between the complexity of models and the quality of their recommendations \citep{ustun_supersparse_2016}.
This is crucial for the successful adoption of decision models in practice, as overly complex models are difficult to analyse by domain experts and impede the manual application by human practitioners. 

Instead of learning standard scoring systems, we are interested in the task of learning probabilistic scoring lists, i.e., in constructing a PSL $h$ from training data 
\begin{equation}\label{eq:td}
\mathcal{D} =  \big\{ (\vec{x}_i , y_i) \big\}_{i=1}^N \subset \mathcal{X} \times \mathcal{Y} \, .
\end{equation}
This essentially means determining the following components:
\begin{itemize}
\item the subset of features to be included and the order of these features;
\item the score assigned to each individual feature;
\item the probabilities for the resulting combinations of stage and total score. 
\end{itemize}
A first question in this regard concerns the quality of a model $h$:
What do we actually mean by a ``good'' probabilistic scoring list?
Intuitively, a good PSL allows for making decisions that are quick and confident at the same time.
Thus, we would like to optimize two criteria simultaneously, namely, to minimize the number of features that need to be determined before a decision is made, and to maximize the confidence of the resulting decision.
This compromise could be formalized in different ways, but regardless of how an overall performance measure is defined, the problem of optimizing that measure over the space of possible PSLs will be computationally hard \citep{chev_rm13}.

\subsection{A Greedy Learning Algorithm}
As a first attempt, we therefore propose a heuristic learning procedure that is somewhat inspired by decision tree learning.
Starting with the empty list, the next feature/score combination $(x_k , s_k)$ is added greedily so as to improve performance the most,\footnote{As the importance of a feature $x_k$, and hence the score $s_k$, can only be decided relative to other features, the choice of the score for the first feature is ambiguous; assuming this feature to be important, we have given it the largest score possible.} and this is continued until no further improvement can be obtained.

To this end, each (remaining) feature/score combination is tried and evaluated as follows: Let $\Sigma_k$ be the set of total scores $T$ in stage $k$ as defined in (\ref{eq:sigmak}), and $Q =\{(N_T, \hat{q}_T) \mid T \in \Sigma_k\}$ the set of probability estimates $\hat{q}_T = \hat{q}(k,T)$ for total scores $T \in \Sigma_k$, together with the number $N_T$ of training examples being assigned this score.
The feature/score combination is then evaluated in terms of the \emph{expected entropy}:
\begin{equation}\label{eq:expected_entropy}
E(Q) = \sum_{(N_T,\hat{q}_T) \in Q} \frac{N_T}{N} \, H \big( \hat{q}_T \big) \, ,    
\end{equation}
where 
$H$ is the Shannon entropy
$$
H(q) = - q \cdot \log(q) - (1-q) \log(1-q) \, .
$$
Thus, according to (\ref{eq:expected_entropy}), the entropy of each distribution $\hat{q}_T$ is weighted by the probability that this distribution occurs.

\subsection{Probability Estimation}
As for the estimation of the probabilities $q(k, T)$, the most obvious idea would be a standard frequentist approach, i.e., to estimate them in terms of relative frequencies $P_T/N_T$, where $N_T$ is again the number of training examples with total score $T$, and $P_T$ is the number of examples with total score $T$ and class $y = 1$ (in stage $k$).
However, as these estimates are obtained independently for each score $T$, they may violate the monotonicity condition (\ref{eq:mono}).
A better idea, therefore, is to estimate them jointly using a probability calibration method \citep{silv_cc21}.
To this end, the original data $\mathcal{D}$, or a subset $\mathcal{D}_{cal}$ specifically reserved for calibration (and not used for training), is first mapped to the data
\begin{equation}\label{eq:cdata}
\mathcal{C}  \defeq  \big\{  (T(\vec{x}), y) \given (\vec{x} , y) \in \mathcal{D}_{cal} \big\} \subset \Sigma_k \times \mathcal{Y} \, ,
\end{equation}
to which any calibration method can then be applied.
One of the most popular techniques, \emph{isotonic regression} \citep{nicu_pg05}, amounts to finding values $\hat{q}(k,T)$ as solutions to the following constrained optimization problem:  
\begin{align*}
& \text{minimize} \quad \sum_{(T,y) \in \mathcal{C}} \big( \hat{q}(k,T) - y \big)^2 \\
& \text{s. t. } \quad \forall\, T, T' \in \Sigma_k: \, (T < T') \Rightarrow (\hat{q}(k,T) \leq \hat{q}(k,T')) 
\end{align*}

\begin{figure}[ht]
	\centering
        \includegraphics[width=1\linewidth]{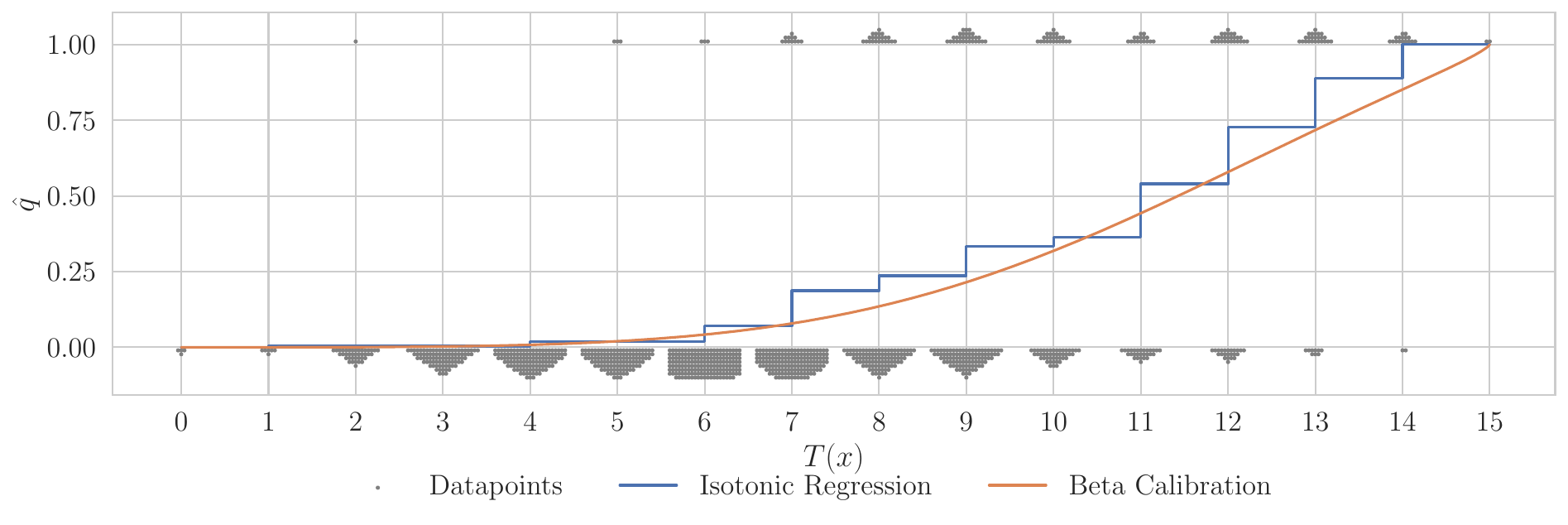}
	\caption{Example of calibration with isotonic regression and beta calibration, using the medical dataset introduced in Section 5. The values on the x-axis correspond to the total scores. As class labels are either 0 or 1, the data points $\mathcal{C}$ are plotted with jittering for better visualization.}
    \label{fig:ir}
\end{figure}

Another common calibration method is the use of a logistic regression, which, however, assumes class-wise normally distributed scores. \citet{kull_bc17} show that this  assumption does not hold for common classifiers and propose the more flexible \emph{beta calibration}, which defines a mapping $[0,1] \fromto [0,1]$ of the following form:
\begin{equation}
\hat{q}(k, \tau ) = \frac{1}{1 + m^a (1 - m)^{-b} \, \tau^{ -a} \, (1 - \tau)^b }   \, ,
\end{equation}
where $a,b \ge 0$ and $m\in [0,1]$ are parameters. 
These parameters are fitted by minimizing log-loss on the calibration data (\ref{eq:cdata}).
In our case, the $\tau$-values are (linear) transformations of the total scores $T$ from $\Sigma_k$ to $[0,1]$.  
Both approaches, isotonic regression and beta calibration, are illustrated in Figure \ref{fig:ir}.


Note that, from a probability estimation point of view, the estimation of one distribution per total score $T \in \Sigma_k$ is a meaningful compromise between a global probability estimate (not taking any context features into account) and a \emph{per-instance} estimation, i.e., the prediction of an individual distribution $\prob( \cdot | \, \vec{x})$ tailored to any specific instance $\vec{x}$.
Obviously, the former is not informative enough, while the latter is very difficult to obtain, due to a lack of statistical information related to a single point \citep{foyg_tl21}.
According to our assumption, all instances with the same total score $T$ share the same probability. Therefore, those instances in the training data with the same score form a homogeneous statistical subgroup
$$
\mathcal{D}_T \defeq \big\{ (\vec{x}_i , y_i) \in \cD \given T(\vec{x}) =  T \big\} \, ,
$$
to which statistical estimation methods can be applied.
While this is in line with other local prediction methods, such as probability estimation trees (PETs) \citep{PETs}, the distinguishing feature here is the way in which the instance space $\mathcal{X}$ is partitioned.
For example, compared to  PETs, PSLs appear to have a more rigid structure, because the succession of tests (features) is fixed and can not vary depending on the value of the features (like in trees).
Moreover, the size of the partition, $|\Sigma_k|$, will normally be smaller than the (up to) $2^k$ different leaf nodes in a tree (leaves with same scores are merged).
Both factors contribute to the increased interpretability of a single sequence of $k$ feature tests, as opposed to up to $2^k$ different paths through a PET.

\subsection{Feature Binarization}
\label{sec:binarization}

The features $\mathcal{F}$ in a PSL are assumed to be binary.
However, in many practical applications, instances are also characterized by continuous attributes. To make scoring systems amenable to these kinds of data, continuous attributes need to be turned into binary features through \emph{binarization}.
This is done by selecting a threshold $t_j$ for each feature $f_j \in \mathcal{F}$ and assigning a value of $0$ for values below the threshold and $1$ otherwise.
Such a discretization from numerical values into binary features is unavoidably accompanied by a loss of information.
To lose as little information as possible about the connection between the feature value and the class label, we employ the \textit{entropy minimization heuristic} \citep{Fayyad1993MultiIntervalDO}.
We can make use of this heuristic either for preparing the data in a preprocessing step (\textit{preprocessing}) or as a subroutine that is used during the greedy construction of the PSL (\textit{in-search}).

When feature binarization is carried out as a preprocessing step, all features are treated independently.
For each (numerical) feature $f_j$, we consider all possible bisections of the dataset when thresholding with $t_j$, where $t_j$ is any mid-point between two consecutive values assumed by that feature in the training data:
\begin{align*}
Y^\le_{t_j} &= \{y_i \mid (\vec{x}_i, y_i) \in \cD, \, \vec{x}_{i,j}\le t_j \}\\
Y^>_{t_j} &= \{y_i \mid (\vec{x}_i, y_i) \in \cD, \, \vec{x}_{i,j}>t_j \}
\end{align*}
All possible bisections are enumerated, and we finally select the threshold that leads to the minimal expected entropy by considering the cardinality and relative frequency of the positive class in each bisection:
$$
t_j^* = \arg\min_{t_j} E\left( \Big\{\Big(|Y^>_{t_j}|, \sum_{y\in Y^>_{t_j}} \frac{y}{|Y^>_{t_j}|}\Big),\Big(|Y^\le_{t_j}|, \sum_{y\in Y^\le_{t_j}} \frac{y}{|Y^\le_{t_j}|}\Big)\Big\}\right)\, ,
$$
with $E$ defined according to (\ref{eq:expected_entropy}).

Instead of finding the optimal threshold $t_j^*$ by ordering all $\vec{x}_{i,j}$ and exhaustively testing every candidate threshold between two consecutive values in a brute-force manner, one can also make use of a heuristic that is computationally less costly.
By assuming quasi-convexity of the expected entropy with respect to the threshold point, hierarchical binary search can be used instead.
In practice, this assumption does not necessarily hold, but is computationally less expensive, as only a logarithmic number of candidate thresholds need to be considered.
In the empirical evaluation in Section \ref{sec:exp}, both the heuristic and brute-force threshold selection are evaluated.

As an alternative to preprocessing, feature binarization can also be carried out \textit{in-search}, i.e., during the proposed greedy learning algorithm.
Here, the features are not treated independently, but rather sequentially.
Recall that when constructing a PSL, the greedy learning algorithm chooses a feature and an associated score in each stage $k$.
Alongside these choices, it now also chooses the binarization threshold for continuous features.
Thus, at construction in stage $k$, the features selected at stages $1,\dots,k-1$ have already been binarized.
Again, binarization is done by enumerating all candidate thresholds $t_k$ and selecting the one that minimizes expected entropy of the resulting partitioning:
$$
t_k^*=\arg\min_{t_k} E\big(\{(N_T, \hat{q}_{t_k}(k,T)) \mid T \in \Sigma_k\}\big) \, ,
$$
with $ \hat{q}_{t_k}(k,T)$ the probability estimate for total score $T$ at stage $k$ when the $k^{th}$ feature is binarized by threshold $t_k$.

\subsection{Beyond Probabilities: Capturing Epistemic Uncertainty}
\label{sec:beyond_probabilities}
Going beyond standard probabilistic prediction, various methods have recently been proposed in machine learning that seek to distinguish between so-called aleatoric and epistemic uncertainty \citep{mpub272,mpub440}.
Broadly speaking, aleatoric uncertainty refers to inherent randomness and stochasticity of the underlying data-generating process.
This type of uncertainty is relevant in our case, because the dependence between total score $T$ and decision/class assignment $y$ is presumably non-deterministic.
Aleatoric uncertainty is properly captured in terms of probabilities, i.e., by the approach introduced above. 

Epistemic uncertainty, on the other side, refers to uncertainty caused by a lack of knowledge, e.g., the learner's uncertainty about the true distribution $p = \prob( \cdot | \, T)$.
In a machine learning context, this uncertainty could be caused by insufficient or low-quality training data. Obviously, it is relevant in our case, too:
Proceeding further in the decision list, the training data will become more and more fragmented, because the number of possible values for the total score increases.
Consequently, the estimation $\hat{q}_T$ of a conditional probability $\prob( y = 1 \given T)$ will be based on fewer and fewer data points, so that the epistemic uncertainty increases (even if the joint estimation of these probabilities for all scores $T$ alleviates this effect to some extent). 

Representing this uncertainty is arguably important from a decision-making point of view.
For example, proceeding in the list and adding another variable may imply that the (predicted) distribution becomes better in the sense of having lower entropy, but at the same time, the prediction itself may become more uncertain. 
In that case, it is not clear whether the current stage should be preferred or maybe the next one\,---\,the answer to this question will depend on the attitude of the decision maker (toward risk), and probably also on the application.

A natural approach to capturing epistemic uncertainty is to replace point estimates $\hat{q}_T$ of $p_T = \prob( y = 1 \given T)$ by interval estimates\,---\,epistemic uncertainty is then reflected by the interval widths.
Formally, we can view the true probability $p_T$ as the (unknown) parameter of a Bernoulli distribution (binomial proportion).
There is vast statistical literature on estimating confidence intervals for binomial proportions, and various constructions of such intervals have been proposed. 
For example, the Clopper-Pearson interval \citep{clopper} with confidence level $1-\alpha$ can be expressed as $l_T \leq p_T \leq u_T$ with
\begin{align*}
l_T & = \left( 1 + \frac{N_T+1}{P_T \, F[\alpha/2; 2P_T, 2(N_T+1)]} \right)^{-1}  \, , \\
u_T & = \left( 1 + \frac{N_T}{(P_T+1) \, F[1-\alpha/2; 2(P_T+1), 2N_T]} \right)^{-1} \, ,
\end{align*}
where $N_T$ is the number of negative examples, $P_T$ the number of positive examples, and $F[c;d,d']$ is the $c$-quantile from an F-distribution with $d$ and $d'$ degrees of freedom. 

On the basis of individual confidence intervals of that kind, a complete \emph{confidence band} 
$\big\{ [l_T^* , u_T^*] \, \vert \, T \in \Sigma_T \big\}$,
i.e., a sequence of intervals for all total score values, can be constructed as follows: 
\begin{itemize}
\item First, one has to guarantee a simultaneous confidence of $1- \alpha$. The simplest way to do so is to apply Bonferroni correction, i.e., to compute individual intervals $[l_T, u_T]$ for confidence level $1- \alpha/|\Sigma_k|$.
\item 
Second, monotonicity constraints can be incorporated by correcting the intervals as follows:
\begin{align}
l_T^* & \leftarrow \max \{ l_V \mid V \in \Sigma_k , \, V \leq T \} \label{eq:corrl}\\
u_T^* & \leftarrow \min \{ u_V \mid V \in \Sigma_k , \, V \geq T \} \label{eq:corru}
\end{align}
\end{itemize}
Note that, although the correction (\ref{eq:corrl}--\ref{eq:corru}) may lead to inconsistencies (empty intervals), it still guarantees the $1-\alpha$ confidence (under the assumption of monotonicity):
With probability (at least) $1-\alpha$, we have $p_T \in [l_T, u_T]$ simultaneously for all $T \in \Sigma_k$, which in turn implies $p_T \geq p_V \geq l_V$ for all $V \leq T$ and $p_T \leq p_V \leq u_V$ for all $V \geq T$.

In order to assure that the confidence band covers the calibrated (point) estimates $\hat{q}(k,T)$, one may consider another correction step (which obviously maintains monotonicity):
\begin{align}
l_T^\star & \leftarrow \min \{ \, \, l_T^* ,  \, \hat{q}(k,T) \, \}  \\
u_T^\star & \leftarrow \max \{ \, u_T^* ,  \, \hat{q}(k,T) \, \}  
\end{align}

\subsection{Ranking}
In addition to the standard probabilistic binary classification task, we also consider PSL for the task of ranking. Instead of a model that assigns each instance to the positive or the negative class, we now seek a model that is able to prioritize instances from most likely positive to most likely negative\,---\,in the literature, this problem is known as the bipartite ranking problem \cite{mpub224}. Again, models of that kind are highly relevant and have many practical applications. 

When having access to standard training data, i.e., a set of instances labelled positive or negative, this can essentially be accomplished using the PSL as is. More specifically, a PSL can be trained in exactly the same way as in the case of binary classification. Then, given a set $X \subset \mathcal{X}$ of instances to be ranked, the PSL can be used to predict a probability $\hat{q}(\vec{x})$ for each $\vec{x} \in X$, and the ranking $\hat{\pi}$ of $X$ is obtained by sorting all $\vec{x} \in X$ in decreasing order of their (predicted) probabilities. Indeed, it can be shown that most common loss functions for bipartite ranking, comparing a predicted ranking $\hat{\pi}$ with a binary ground-truth, are minimized (in expectation) by sorting instances in decreasing order of their probability of being positive \cite{mpub224}. An important example of such a loss is the well-known AUC measure\footnote{Actually, AUC in an accuracy measure, so the loss would be obtained by $1-$AUC.} \citep{Fawcett06}.

However, in the realm of ranking, training data is often given in the form of relative comparisons $\vec{x} \succ \vec{x'}$ between instances $\vec{x},\vec{x'} \in \mathcal{X}$, suggesting that $\vec{x}$ is preferred to (should be ranked higher than) $\vec{x}'$.
Then, in contrast to the first scenario, the PSL algorithm cannot be applied in a straightforward way. In particular, the expected entropy (\ref{eq:expected_entropy}) cannot be computed as an impurity measure and selection criterion in the greedy learning algorithm. Instead, one has to refer to a ranking loss. 

Given a set of pairwise comparisons $\mathcal{D} \defeq \{ \vec{x}_i \succ \vec{x}'_i \}_{i=1}^N$, the \emph{pairwise soft rank loss} is defined as follows:
\begin{equation}
\label{eq:rank_loss}
 \text{SRL}(\mathcal{D}) \defeq \frac{1}{N} \sum_{i=1}^N  \ell (\vec{x}_i,\vec{x}'_i)
 \end{equation}
with 
$$ \ell (\vec{x},\vec{x'}) =
\begin{cases}
    0   & \text{if } \hat{q}(\vec{x}) > \hat{q}(\vec{x}') \\
    0.5 & \text{if } \hat{q}(\vec{x}) = \hat{q}(\vec{x}') \\
    1   & \text{if } \hat{q}(\vec{x}) < \hat{q}(\vec{x}') \\
\end{cases} \, .
$$
\smallskip

\noindent
This loss imposes a penalty of $0$ for pairs that are correctly ordered, $0.5$ for ties, and $1$ for incorrect orderings.
It can be computed in each stage $k$ of the PSL algorithm\,---\,the probabilities $\hat{q}(\vec{x})$ are then given by $\hat{q}\big(k,T(\vec{x})\big)$. 
The PSL algorithm itself is then modified in the sense that, in every stage $k$, it finds the feature/score combination that yields the smallest SRL instead of the smallest expected entropy. 

The PSL has an interesting and intuitive interpretation in the context of ranking: Starting with the entire set $X$ as a single tie group, it successively refines a ranking by splitting such groups into smaller subgroups and sorting these subgroups.
In the first stage, all instances are assigned the same probability, i.e., there are only ties, and we start with an AUC of $0.5$.
As we progress throughout the stages, the set of total scores becomes larger, and more and more ties are being resolved.
The process can then be stopped as soon as a sufficient resolution has been reached. 
This is particularly useful in scenarios in which one is not interested in retrieving the complete ordering of alternatives, but rather in eliciting the top (or bottom) $m$ alternatives.

\section{Empirical Evaluation}
\label{sec:exp}

In this section, we present a case study in medical decision-making meant as a first evaluation of our approach.
Unless stated differently, the figures show the models' mean performance and a 95\% confidence interval of the mean, aggregated over 100 Monte Carlo cross-validation (MCCV) splits with $2/3$ used for training and $1/3$ used in testing.
Additionally, the score set 
$\{ \pm 1, \pm 2, \pm 3\}$
was chosen.
All experiments in this section were executed on an Intel i7-9750H in less than one hour.
The detailed experimental setup and implementation is publicly available\footnote{\url{https://github.com/TRR318/pub-ml-psl/releases/tag/v1.0.0}} as is the implementation of the learning algorithm.\footnote{\url{https://github.com/TRR318/scikit-psl/releases/tag/v0.7.0}}

In the next sections, we will attempt to answer the following research questions:
\begin{description}
    \item[RQ1:] Is greedy search sufficient to find a good model?   
    \item[RQ2:] Are the probability estimates of the PSL well calibrated?
    \item[RQ3:] How do the Clopper-Pearson confidence intervals reflect the increase in epistemic uncertainty throughout the PSL stages?
    \item[RQ4:] How does binarization as preprocessing compare with in-search binarization? Do the heuristic simplifications in the binarization hamper PSL performance?
    \item[RQ5:] Is the PSL applicable for ranking tasks? How well does it perform when being trained on class labels versus preference data? 
\end{description}

First, however, we briefly summarize the datasets used in this study.

\subsection{Datasets}
\paragraph{Coronary Heart Disease Data}

The dataset for this case study has originally been used to evaluate the diagnostic accuracy of symptoms and signs for coronary heart disease (CHD) in patients presenting with chest pain in primary care.
Chest pain is a common complaint in primary care, with CHD being the most concerning of many potential causes.
Based on the medical history and physical examination, general practitioners (GPs) have to classify patients into two classes:
patients in whom an underlying CHD can be safely ruled out (the negative class) and patients in whom chest pain is probably caused by CHD (the positive class). 

Briefly, 74 general practitioners (GP) recruited consecutively patients aged $\geq 35$ who presented with chest pain as primary or secondary complaint.
GPs took a standardized history and performed a physical examination.
Patients and GPs were contacted six weeks and six months after the consultation.
All relevant information about course of chest pain, diagnostic procedures and treatments had been gathered during six months.
An independent expert panel of one cardiologist, one GP and one research staff member reviewed each patient's data and established the reference diagnosis by deciding whether CHD was the underlying reason of chest pain. For details about the design and conduct of the study, we refer to \citet{boes_a010}.

Overall, the dataset comprises 1199 (135 CHD and 1064 non-CHD) patients described by ten binary attributes: 
($f_1$) patient assumes pain is of cardiac origin, ($f_2$) muscle tension, ($f_3$) age gender compound, ($f_4$) pain is sharp, ($f_5$) pain depends on exercise, ($f_6$) known clinical vascular disease, ($f_7$) diabetes, ($f_8$) heart failure, ($f_9$) pain is not reproducible by palpation, ($f_{10}$) patient has cough.
Note that, by way of domain knowledge, all these features can be encoded in such a way that the presence of a feature does always increase the likelihood of the positive class.
Therefore, scoring systems can be restricted to positive scores.
For the following experiments, the missing feature values have been imputed using the mode, representing the most frequent value of each feature.

\paragraph{UCI Datasets}
In addition to the CHD dataset, we also consider two datasets from the UCI repository \citep{misc_breast_cancer_coimbra_451, misc_ilpd_(indian_liver_patient_dataset)_225}. 
The first is concerned with the prediction of \emph{breast cancer} for patients in Coimbra, Portugal.
It comprises 9 features including resistin, glucose, age, and BMI.
The dataset is of small size, with only 116 instances, of which 64 are positive. 
The second UCI dataset is concerned with \emph{Indian liver patients} (ILP). 
It contains 583 instances, of which 416 are positive, i.e., the majority of patients are positive.
The dataset contains 10 features including age, sex, total proteins.
In contrast to the CHD data, the UCI datasets contain numeric features, which have to be binarized, e.g., by using the methods described in Section~\ref{sec:binarization}.

\subsection{RQ1: Expected Entropy Minimization}
\begin{figure}[htb]
	            \includegraphics[width=\linewidth]{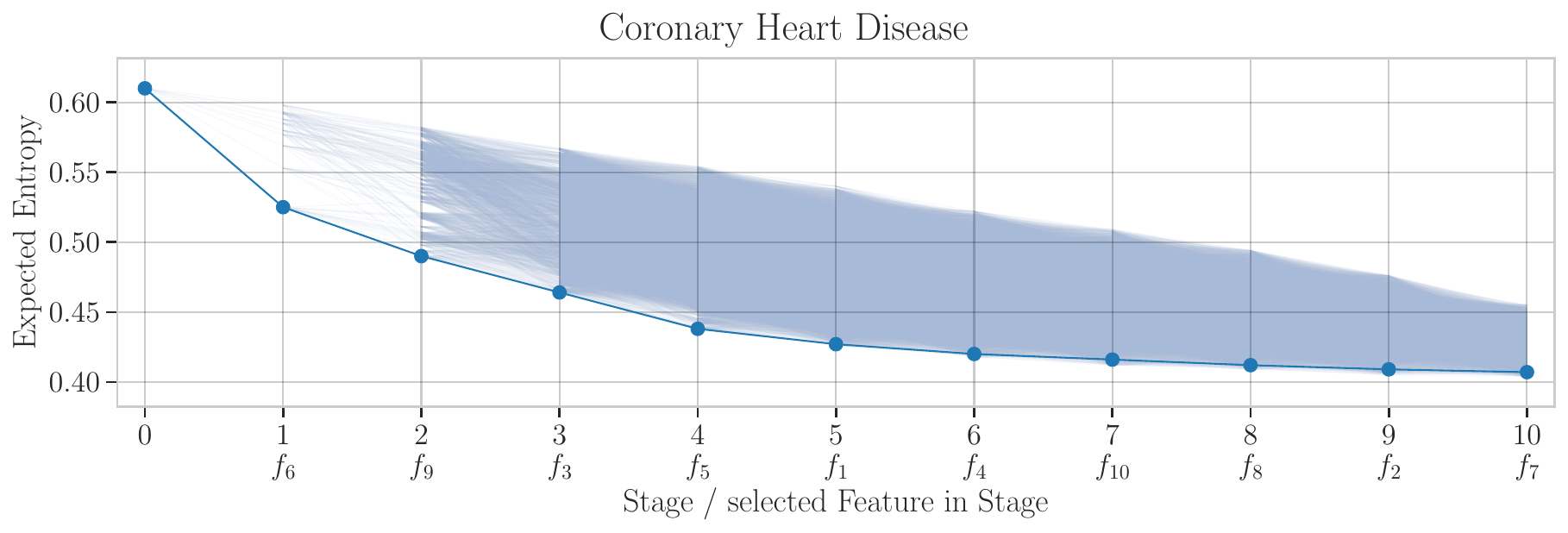}
	\caption{Evaluation of the greedy learning algorithm (blue line) on the coronary heart disease dataset. The light blue lines show the complete search space induced by all feature permutations and possible score assignments.
 The features, selected by the greedy algorithm in every stage, are also labelled on the x-axis.
 The visualization was created for a score set $\mathcal{S} = \{1,2,3\}$.}
    \label{fig:greedy_evaluation}
\bigskip
        \includegraphics[width=\linewidth]{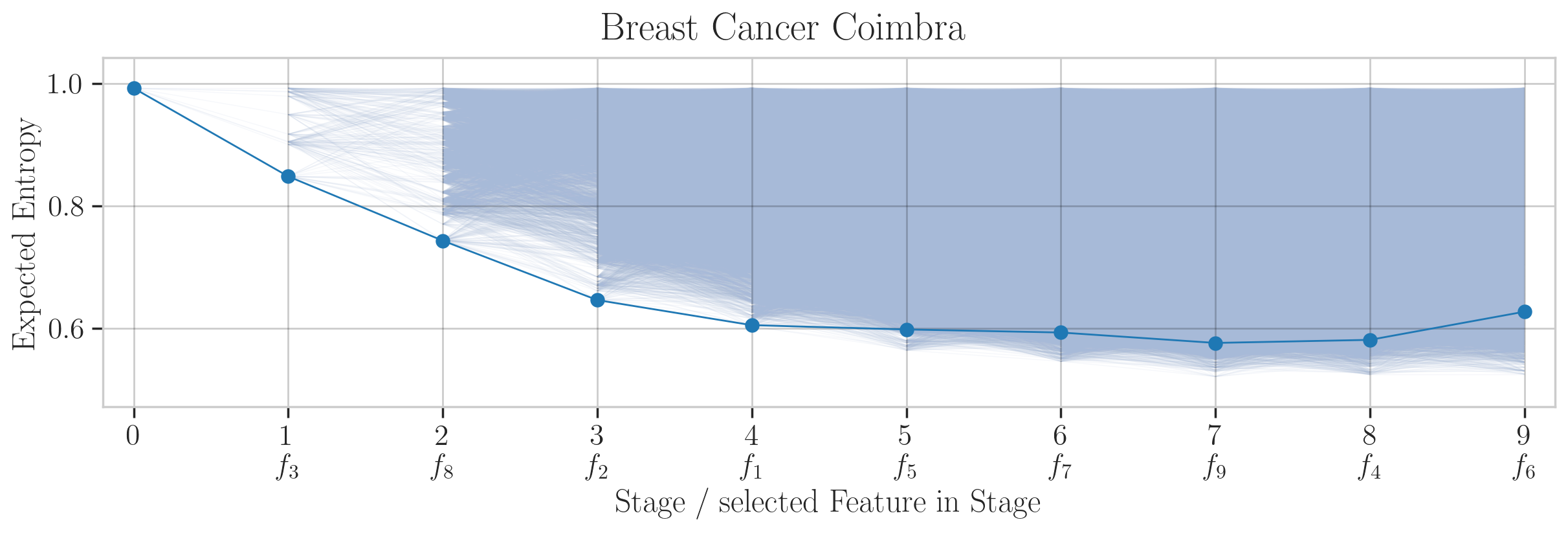}
\caption{Evaluation of the greedy algorithm on the breast cancer Coimbra dataset with a score set of
  $\mathcal{S} = \{-2,-1,+1,+2\}$ 
  (refer to Figure~\ref{fig:greedy_evaluation} for details).}
     \label{fig:greedy_evaluation_combria}
\end{figure}

The introduced algorithm iteratively selects the feature/score pair that minimizes the expected entropy \eqref{eq:expected_entropy} for each stage.
As can be seen in Figure \ref{fig:greedy_evaluation}, entropy continues to decrease, but the improvements diminish stage by stage and almost vanish after the addition of the fifth feature.
Interestingly, this result is very much in agreement with previous studies on this data, and the top-5 features in  Figure~\ref{fig:greedy_evaluation} exactly correspond to those features that have eventually been included in the ``Marburg Heart Score'',  a decision rule that is now in practical use.\footnote{\url{https://www.mdcalc.com/calc/4022/marburg-heart-score-mhs}}

As our algorithm minimizes expected entropy on the training data greedily, one may wonder to what extent expected entropy is also minimized globally, i.e., across all stages.
To get an idea, we compared the expected entropy curve produced by the greedy algorithm with the curves produced by all other PSLs.
With the score set $\mathcal{S} = \{1,2,3\}$\,---\,a complete enumeration of the resulting set of PSLs is still feasible. 
As can be seen from Figure~\ref{fig:greedy_evaluation}, the greedy approach (shown in solid blue) performs well, at least on the CHD dataset.
Figure~\ref{fig:greedy_evaluation_combria} illustrates the result of the greedy parameter search on the BCC dataset.
As we can see here, the curve of the selected model is clearly not the lower envelope, i.e. it does not achieve stagewise optimal performance.
This may be due to the greedy search approach, which acts myopically in the sense that it irrevocably selects the locally best option and therefore could be missing out on global optima.
However, note that stagewise optimal performance may also be impossible to achieve with a single scoring list, as the stagewise best performing models may stem from separate lists having different prefixes of feature/score pairs.

\subsection{RQ2: Investigating Probability Estimates}
\label{sec:investigating_prob_est}
Next, we investigate the trustworthiness of the probability (point) estimates of our proposed method. 
As already outlined in Section \ref{sec:beyond_probabilities}, the training data becomes more and more fragmented when progressing in the PSL, as the set of possible total scores grows.
Consequently, the probability estimates for the individual instances are based on fewer and fewer training data.
While the final probabilistic predictions are based on a joint estimation in terms of probability calibration, it is not clear how the quality of the estimations develops throughout the PSL stages. 

\begin{figure}[ht!]
    \centering
    \includegraphics[width=\linewidth]{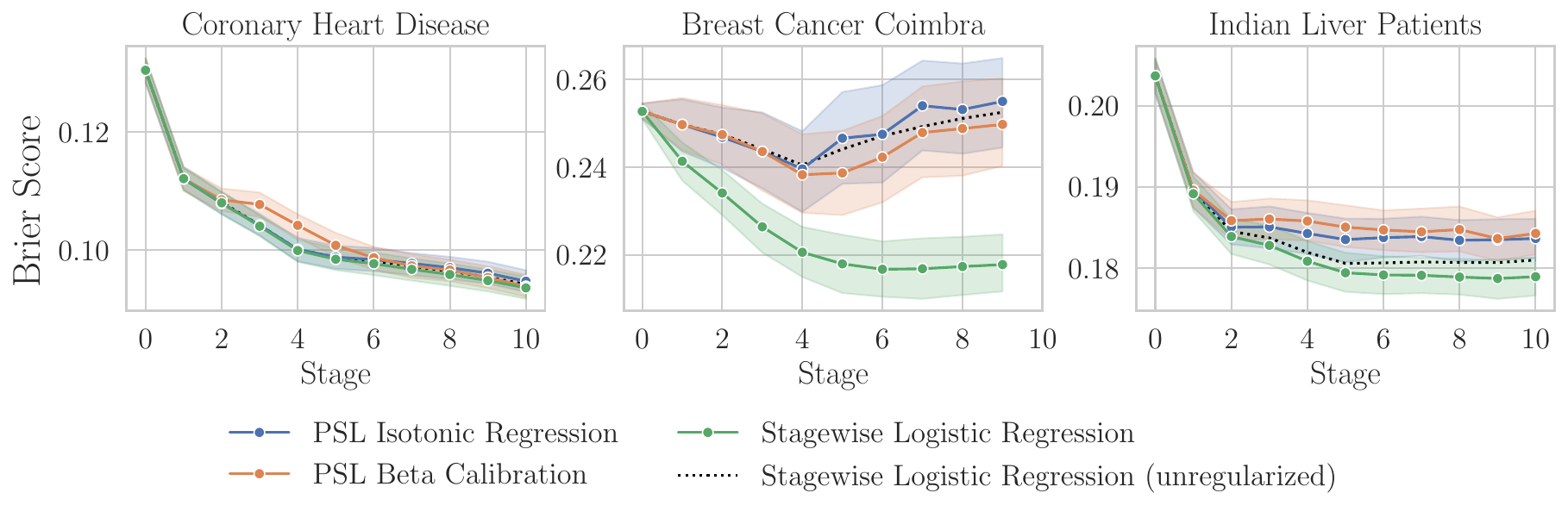}
    \caption{Stagewise Brier Score for PSLs on the test datasets.}
    \label{fig:stagewise_brier}
\end{figure}

We evaluate this quality by computing the stagewise \textit{Brier score} \citep{brier1950verification}.
Let $T_k(\vec{x}_i)$ denote the total score of some instance $\vec{x}_i$ at stage $k$.
Having access to a set of test data $\mathcal{D}_{test} \defeq \{(\vec{x}_i , y_i)\}_{i=1}^N$, the Brier score at stage $k$ is given by 
$$BS(k) \defeq \frac{1}{N} \sum_{i=1}^N \big(\hat{q}(k,T_k(\vec{x}_i)) - y_i \big)^2.$$ 
Figure \ref{fig:stagewise_brier} shows the stagewise Brier score for all considered datasets when using isotonic regression or beta calibration for the PSL. 
As a baseline, we also train a logistic regression (LR) model, using the same features as PSL in each stage. Note that, compared to PSL, LR is more flexible in the sense that scores are real-valued and not restricted to (small) integers.
Additionally, the learning algorithm of logistic regression employs $L_2$-regularization in order to prevent the model from overfitting.
For all approaches, the features were binarized in advance (\textit{preprocessing}) using the bisect heuristic (cf.\ Section~\ref{sec:binarization}).

As we can see, isotonic regression and beta calibration perform quite similarly for each of the considered datasets.
On the CHD dataset, both are also on par with the stagewise logistic regression, which is known to produce well-calibrated models.
However, on the BCC as well as the ILP dataset, the PSL variants achieve worse Brier scores than the logistic regression baseline. 
Their Brier scores even increase for higher stages, which is likely to be an overfitting effect:
The dotted line indicates the performance of an unregularized LR, whose performance is very similar to the two considered PSL variants.
Recall that the BCC dataset only contains 116 instances.  
The regularized LR manages to avoid overfitting the training data and achieves better generalization performance in terms of stagewise Brier score.
Although the PSL is a quite restricted model with only a few integer weights, it tends to overfit if there is not enough training data available, e.g., in the case of the BCC dataset. 
Thus, when relying on point estimates and having only access to small amounts of training data, PSL needs to be regularized to circumvent this problem.
Another alternative is to go from point to interval estimates, e.g., by using the Clopper-Pearson confidence intervals introduced in Section~\ref{sec:beyond_probabilities}, whose application will be examined in the following section.

\subsection{RQ3: Uncertainty Quantification and Decision-Making}
In the previous section, we have discussed that the probability estimates in higher stages of the model are based on less and less data points as they get more fragmented.
In this section, we investigate the applicability of the Clopper–Pearson confidence interval introduced in Section \ref{sec:beyond_probabilities} in medical decision-making.

\begin{figure}[b]
    \centering
    \includegraphics[width=1\linewidth]{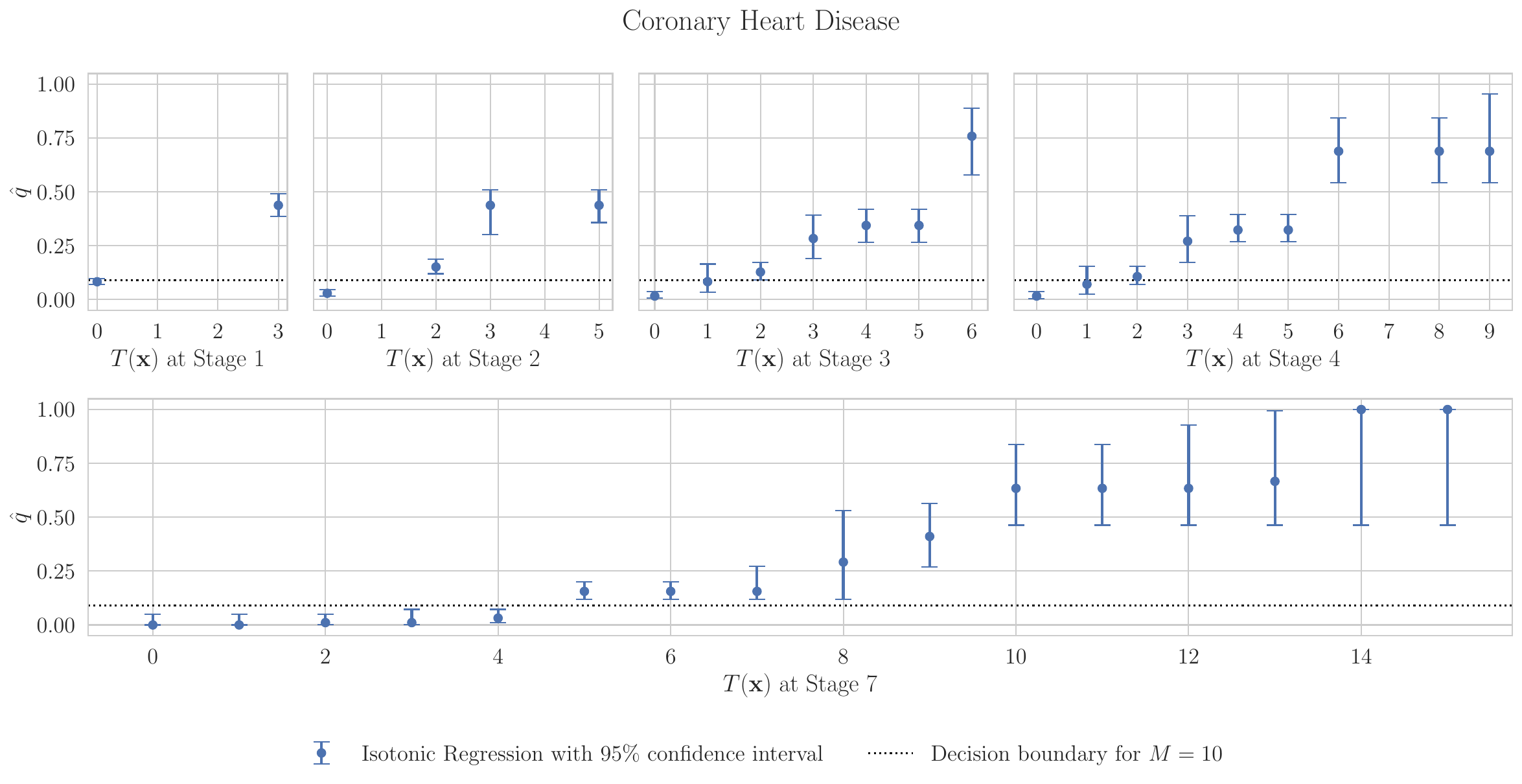}
    \caption{Probability estimates of all possible total scores for the first 4 stages and stage 7 of the PSL, trained on the full CHD dataset. The error bars show the 95\% confidence interval described in Section~\ref{sec:beyond_probabilities}.}
    \label{fig:ci_of_proba}
\end{figure}

Figure \ref{fig:ci_of_proba} illustrates the point estimates as well as the confidence intervals exemplarily for several stages of a PSL trained on the CHD data. 
As expected, we observe an increased size of the confidence intervals in higher stages.
This is not only caused by the increasingly fewer data points the relative frequency estimate is based on, but also by the Bonferroni correction, that grows in the size of the set of total scores $|\Sigma_k|$.
For example, in stage $4$, there is not a single data point that exhibits a total score of $7$, hence the confidence bounds are fully determined by the neighbouring total scores.

Since the CHD dataset is quite imbalanced, there are fewer data points for positive samples, hence the confidence intervals for predicting high true-class probabilities are wider, as there are fewer data points for the respective total scores.
For stage 6 and beyond (see Figure \ref{fig:ci_of_proba}) the lower confidence bound never exceeds 0.5, meaning that the positive class can never be predicted with high confidence.
Note, that the Clopper-Pearson confidence interval is a conservative guarantee of the probability estimate, i.e., when computing an $95\%$ interval, the probability of the true parameter $p_T$ laying outside the interval is \textit{at most} $5\%$.

Following this illustration of the Clopper-Pearson confidence intervals, we will showcase their usefulness in the context of risk-averse decision-making in medicine.
In medical diagnosis, the consequences of a false negative prediction, i.e., not treating an ill patient, are typically far more severe than of a false positive.
This asymmetry can be captured by a loss function that assigns a loss of 1 to a false positive and a loss of $M \gg 1$ to a false negative.
In the medical domain, this also goes under the notion of ``diagnostic regret'', and various empirical methods for eliciting preferences in decision-making (i.e., the cost factor $M$) have been proposed in the literature \citep{tsal_ar10,more_wh09}. 

To minimize the risk of the decision, the negative class should only be predicted if its probability ($1- \hat{p}$) is $M$ times as high as the probability for the positive $\hat{p}$ class:
$$
\hat{y} = \left\{ \begin{array}{cl}
1 & \text{ if } 1-\hat{p} < M \cdot \hat{p} \\
0 & \text{ otherwise} 
\end{array} \right.
$$
The (estimated) expected loss for this risk-minimizing decision is therefore $E(\hat{y}) = \min \{ 1-\hat{p} , M \cdot \hat{p} \}$.
This decision boundary for $M=10$ is visualized in Figure \ref{fig:ci_of_proba}.
This decision strategy nicely emphasizes the importance of (accurate) probabilistic predictions and, more generally, uncertainty-awareness, in safety-critical domains.

To incorporate risk-awareness into the decision-making process, we propose to select not with respect to the point estimate of the probability, but the upper confidence bound.
Again, referring to Figure \ref{fig:ci_of_proba}, we can observe that deciding with respect to the probability point estimate, data points with score 0 at stage 0 are classified negatively as they lay below the decision boundary.
When using the upper confidence bound instead, all data points are classified negatively, as the upper confidence bound is slightly above the decision threshold.

Again, we compare the PSL variants to an LR model that is trained on the same features as the PSL on each stage.
Accounting for uncertainty, we do not use the point estimate for the predicting the positive class but rather the upper bound of the corresponding $50\%$ Clopper-Pearson confidence interval of the PSL estimate.
\begin{figure}[b!]
	\centering
        \includegraphics[width=\linewidth]{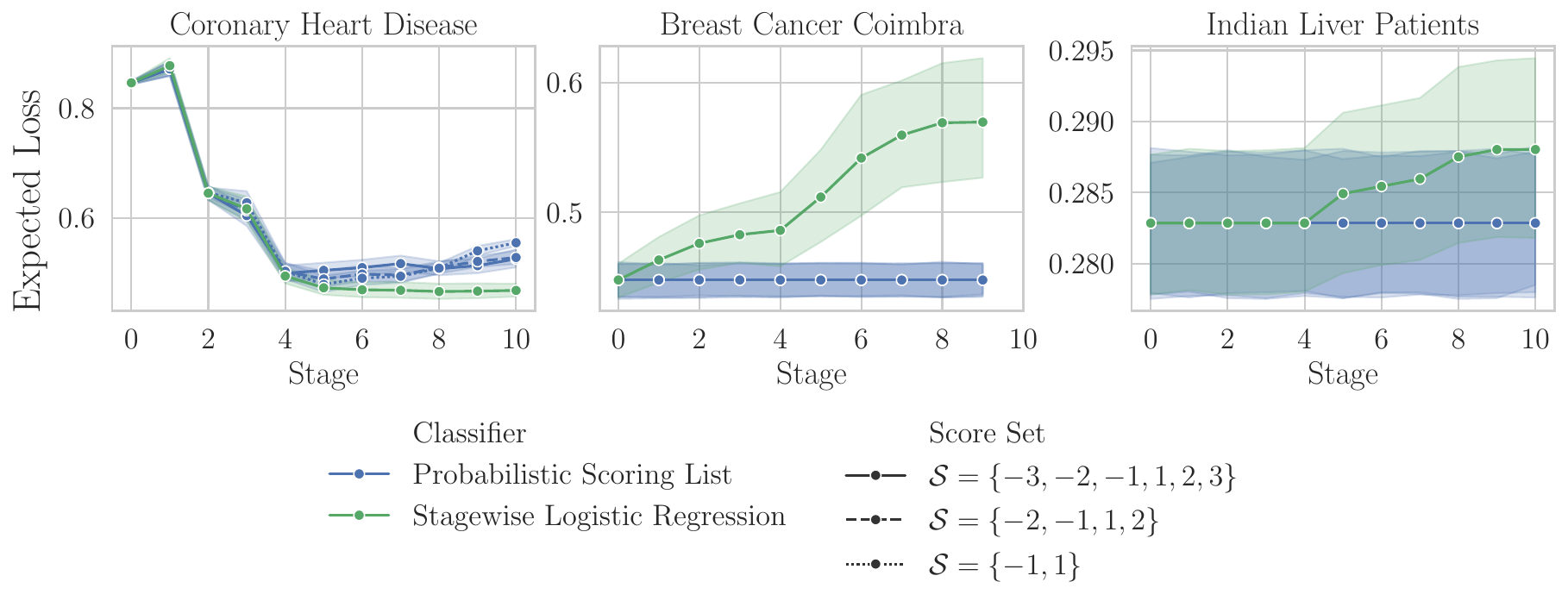}
	\caption{Expected loss, calculated using the upper confidence bound of the 50\% confidence interval.}
    \label{fig:plot_expected_loss}
\end{figure}
Figure \ref{fig:plot_expected_loss} shows the loss for $M=10$.
The PSL has been configured with three different score sets.
For the CHD dataset, we can see that all three PSL variants perform quite similarly, with small improvements for larger score sets.
Moreover, they are all on a par with LR, sometimes even a bit better, which is quite remarkable. 
For all variants, we observe a monotonic decrease in loss until the fifth feature is added.
Again, in the large majority of cases, the five top-features correspond to the features also included in the Marburg heart score.
Adding further features leads to a slight deterioration for PSL.

As more features increase the capacity of the learner, this may look like a standard overfitting effect. However, there is also an alternative, in a sense even opposite explanation.  
Note that we do not observe this deterioration for LR, which can modulate the influence of any additional feature in a very flexible way, by appropriately tuning the weight coefficient\,---\,up to completely ignoring a presumably unuseful feature by setting its weight to 0. PSL does not have this ability. Instead, it can only weight all features in (more or less) the same way.
Therefore, in cases where adding another feature might be useful, but with a weight much smaller than the others, it might be better to omit it completely instead of giving it the same influence as the more important features. Seen from this perspective, the class of scoring systems is simply not flexible enough, and the deterioration might be due to a problem of underfitting rather than overfitting. 

For the BCC and the ILP dataset, the plots look drastically different.
Recall, that at stage $0$, the probability estimate is simply the relative frequency of the positive class in the training data.
Thus, all instances are classified identically.
Employing the risk-averse decision rule, we classify all patients as positive.
Advancing from this is only possible if we make true negative predictions.
For each false negative prediction, we need to make at least $M=10$ true negative predictions, otherwise the expected loss will increase.
Depending on the classification task at hand, this may be a very difficult problem.
We observe that the PSL with $50\%$ confidence intervals refrains from changing the initial classification, which results in a constant value of expected loss throughout the stages.
LR on the other hand acts less risk-averse and introduces negative predictions.
However, it does not introduce enough true negative predictions in order to compensate for the false negatives, resulting in a deterioration of the expected loss.

\subsection{RQ4: Binarization}
As described in Section \ref{sec:binarization}, dealing with numerical features is of great practical importance.
We compare the \textit{in-search} binarization of numerical features with the case in which the features are binarized independently in a \textit{preprocessing} step.
Additionally, we evaluate our heuristic optimization method, with which only a logarithmic number of candidate binarization thresholds need to be checked.

\begin{figure}[b]
    \centering
    \includegraphics[width=\linewidth]{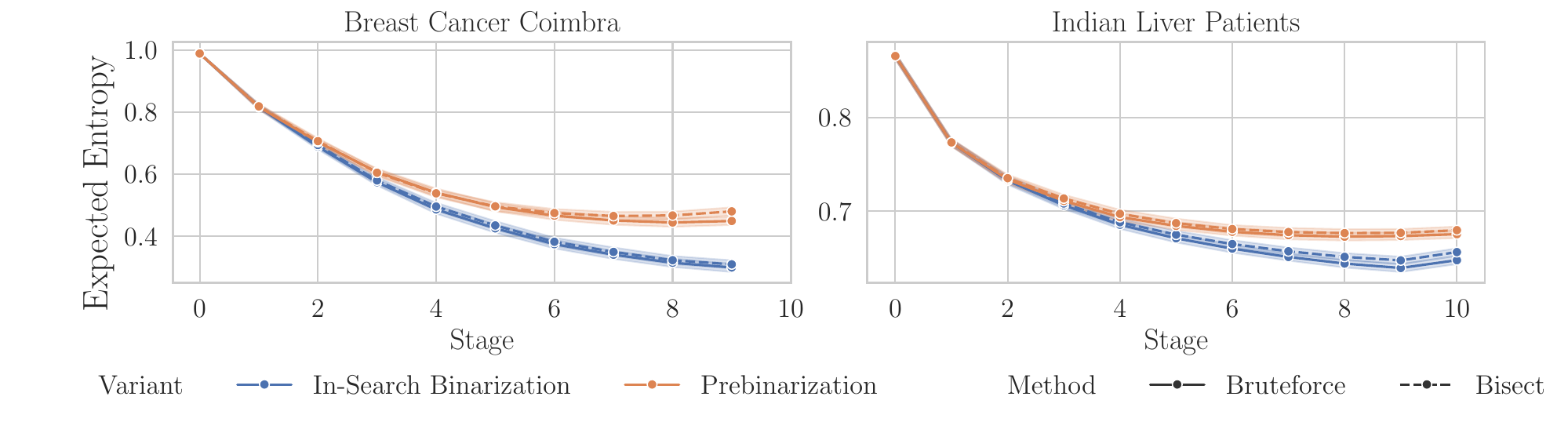}
    \caption{Comparison of pre-binarization and in-search binarization, with heuristic and brute-force threshold optimization.}
    \label{fig:binarization_comparison}
\end{figure}

Figure \ref{fig:binarization_comparison} shows the comparison on the \textit{Breast Cancer} and the \textit{Indian Liver Patient} datasets in terms of the stagewise expected entropy.
In both datasets, the \textit{in-search} binarization is advantageous.
The advantage becomes more pronounced throughout the stages of the PSL.
Intuitively, this can be explained by the fact that in \textit{preprocessing}, the features are binarized individually.
In contrast to that, the \textit{in-search} procedure binarizes the features sequentially during the greedy construction of the PSL.
Thus, when adding a feature/score pair to the PSL, it can take into account dependencies between already selected features and the newly chosen feature by setting the binarization threshold accordingly.
As discussed previously, in situations where adding another feature with the same weight as previous (more important) features is undesirable, abstaining from using a feature is a sensible option and avoids performance deterioration.
\textit{In-search} binarization can enable this, by setting the threshold higher than the maximum value of the feature, effectively ignoring it even if a non-zero score is assigned.

For both methods, we observe that the bisecting search heuristic leads to a negligible deterioration of performance in terms of expected entropy.
Consequently, we consider \textit{in-search} binarization with the bisect search a sensible default configuration and use it for the remaining experiments of this paper.

\subsection{RQ5: Ranking}
In the following, we will evaluate the ranking performance of our proposed method.
To this end, we consider the AUC (Area under the ROC Curve) \citep{Fawcett06}.
For a binary scoring classifier, the AUC can be interpreted as the probability of ranking a randomly chosen positive example before a randomly chosen negative example.

When fitting the classifier, we consider two settings:
In the first one, we assume to have a training dataset as described in the previous experiments of this paper.
Here, we use the standard greedy learning algorithm optimizing the expected entropy \eqref{eq:expected_entropy}.
In the second scenario, we exclusively require access to pairwise comparisons $x \succ x'$ with $x,x' \in \mathcal{X}$ and instantiate the PSL with the pairwise soft rank loss \eqref{eq:rank_loss}.

\begin{figure}[h]
    \centering
    \includegraphics[width=\linewidth]{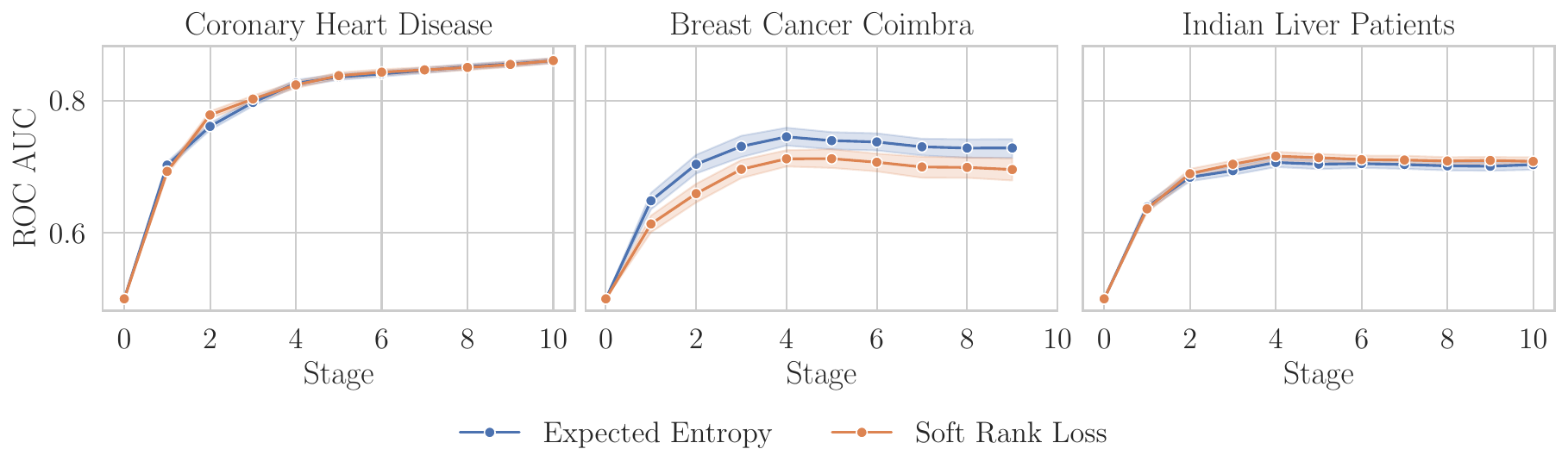}
    \caption{Stagewise AUC (Area under the ROC Curve) for all datasets.}
    \label{fig:roc_auc}
\end{figure}

Figure \ref{fig:roc_auc} shows the stagewise AUC.
For the CHD dataset, we observe a monotonic increase for both settings.
For the other two datasets, the values increase in the beginning until a plateau is reached at stage 4.
The two settings lead to very similar results for the BCC and ILP datasets.
For the breast cancer dataset, optimizing expected entropy leads to better AUC values than the soft rank loss.
However, as already mentioned, the BCC dataset is very small.
We conclude that the PSL is generally applicable for the task of ranking.

\section{Summary and Conclusion}
\label{sec:future_work}
In this paper, we introduced probabilistic scoring lists, a probabilistic extension of scoring systems. Their main advantage and intended use is that they not only allow one to obtain probability estimates that correspond to the scores of the underlying scoring system, but that these estimates can be gradually refined by adding more features. This may, e.g., be important if features are expensive or time-consuming to obtain, so that rough estimates can be obtained cheaply and quickly, and be further refined once additional evidence comes in. In particular, it also allows one to end the decision-making process once a certain probability threshold has been surpassed, thereby allowing a dynamic adjustment of the number of features needed for a positive or negative decision.

Building on the approach presented in this paper, we plan to address the following extensions in future work:
\begin{itemize}
    \item Although the greedy learning algorithm proposed in this paper seems to perform quite well, more sophisticated algorithms for learning PSLs should be developed, including algorithms tailored to specific loss functions.
    \item So far, we only considered the case of binary decisions, which is common for scoring systems; yet, an extension to decision spaces of higher cardinality (polychotomous classification) is practically relevant.

    \item The Clopper-Pearson confidence intervals introduced in Section \ref{sec:beyond_probabilities} are rather loose and for large $\Sigma_k$ (induced by many features or large score sets) they quickly become dominated by Bonferroni correction.
    Thus, we are interested in investigating ways for computing tighter confidence bounds.
    \item So far, we implicitly considered features with unit cost.
    In practice, the cost of feature acquisition may vary drastically for different features.
    For example, measuring the temperature of a patient can be done much more easily than a sophisticated blood test.
    An interesting direction for future work is to consider feature costs explicitly and develop a learning algorithm for cost-effective PSLs \citep{cler_ic19}.
    \item In practice, some features may be not available at inference time, e.g., because there is no access to laboratory equipment in the field.
    Here, it would be interesting to consider default scores for missing features and develop a more sensitive version of the PSL that produces models that are robust to missing features.
    \item As discussed in Section \ref{sec:investigating_prob_est}, we observe an overfitting effect with respect to the probabilities estimated by PSL when only small amounts of training data are available.
    Thus, incorporating regularization in the training procedure of PSLs is an important next step.
    \item So far, we have only examined the PSL from an algorithmic point of view.
    Conducting a user study and investigating how suitable it is for the task of supporting human decision makers is of major interest.
    This also involves the development of usability features such as advanced visualizations of PSL stages.
\end{itemize}

\section*{Acknowledgments}
We gratefully acknowledge funding by the German Research Foundation  (Deutsche Forschungsgemeinschaft, DFG): TRR 318/1 2021 -- 438445824 and the German Research Foundation (DFG) within the Collaborative Research Center ``On-The-Fly Computing'' (SFB 901/3 project no.\ 160364472).

\bibliography{sn-bibliography}

\end{document}